\documentclass[10pt,english]{article}
\usepackage{subfigure}
\usepackage{graphicx}
\usepackage{float}
\usepackage[T1]{fontenc}
\usepackage[latin9]{inputenc}
\usepackage[margin=1.25in]{geometry}
\usepackage{float}
\usepackage{amsmath}
\usepackage{amsbsy}
\usepackage{setspace}
\usepackage{amssymb}
\usepackage{esint}
\usepackage{cite}
\newfloat{algorithm}{tbp}{loa}
\floatname{algorithm}{Algorithm}
\usepackage{hyperref}
\usepackage{listings}

\newlength\myindent
\makeatletter

\floatstyle{ruled}
\newfloat{algorithm}{tbp}{loa}
\floatname{algorithm}{Algorithm}

\usepackage{babel}

\begin{document}

\title{ELM Ridge Regression Boosting}

\author{M. Andrecut}

\date{October 24, 2023}

\maketitle
{

\centering Unlimited Analytics Inc.

\centering Calgary, Alberta, Canada

\centering mircea.andrecut@gmail.com

} 
\begin{abstract}
We discuss a boosting approach for the Ridge Regression (RR) method, with applications to the Extreme Learning Machine (ELM), and we show that 
the proposed method significantly improves the classification performance and robustness of ELMs. 
\end{abstract}

\section{Introduction}

In this short note we consider a class of simple feed-forward neural networks \cite{key-1}, also known as Extreme Learning Machines (ELM) \cite{key-2}. ELMs consist of a hidden layer where the data 
is encoded using random projections, and an output layer where the weights are computed using the Ridge Regression (RR) method. 
Here we propose a new RR boosting approach for ELMs, which significantly improves their classification performance and robustness. 

Let us assume that $X= [x_{n,m}]_{N\times M}$  is a data matrix, where each row $x_n=[x_{n,0},x_{n,1},...x_{n,M-1}] \in \mathbb{R}^M$ is a data point 
from one of the $K$ classes. The classification problem requires the mapping of the rows of a new, unclassified data matrix $\tilde{X}=[\tilde{x}_{nm}]_{N'\times M}$, to the corresponding classes $\{0,1,...,K-1\}$. 
The first layer of the ELM encodes both $X$ and $\tilde{X}$ matrices using the same random projections matrix $R= [x_{j,m}]_{J\times M}$ drawn from the normal distribution $r_{j,m}\in \mathcal{N}(0,1)$:
\begin{equation}
H = h(XR^T), \quad \tilde{H} = h(\tilde{X}R^T), 
\end{equation}
where $h()$ is the activation function, which is applied applied element-wise. 
The second layer of ELM solves the Ridge Regression (RR) problem:
\begin{equation}
W = \text{arg}\min_W \Vert HW - Y \Vert^2 + \lambda \Vert W \Vert^2,
\end{equation}
where $Y=[y_{nk}]_{N\times K}$ is the $N\times K$ target matrix for $K$ classes, and $\lambda > 0$ is the regularization parameter. 

Each row $y_n \in \mathbb{R}^K$ of $Y$ corresponds to the class of the data point $x_n$. The classes are encoded using the one-hot encoding approach: 
\begin{equation}
x_n \in C_k \Leftrightarrow y_{ni} = 
  \begin{cases}
   1  & \text{if } i = k \\
   0 & \text{otherwise }
  \end{cases}.
\end{equation}
The solution of the above RR problem is:
\begin{equation}
W = (HH^T + \lambda I)^{-1}H^TY,
\end{equation}
where $I$ is the $J\times J$ identity matrix. 

Therefore, in order to classify the rows of a new data matrix $\tilde{X}$ we use the following criterion: 
\begin{equation}
\tilde{x}_n \in C_k \quad \text{if} \quad k = \text{arg}\max_{k=0.1,...,K-1} \tilde{y}_{nk}, \quad n=0,1,...,N'-1,
\end{equation}
where
\begin{equation}
\tilde{Y} = [\tilde{y}_{n,k}]_{N' \times K} = \tilde{H}W.
\end{equation}

\section{Boosting method}

Several boosting methods have been previously proposed for the RR problem \cite{key-3}, \cite{key-4}, \cite{key-5}. Our approach here is 
different, and it uses several levels of boosting.  

At the first boosting level, $\ell=0$, one computes the approximation:
\begin{equation}
W_0^0 = \text{arg}\min_{W_0^0}  \Vert H_0^0W_0^0 -  Y  \Vert^2 + \lambda \Vert W_0^0 \Vert^2,
\end{equation}
\begin{equation}
W_0^0 = (H_0^{0T}H_0^0 + \lambda I)^{-1}H_0^{0T}Y,
\end{equation}
and then continues by successively solving for the next $T-1$ approximations:
\begin{equation}
W_{t}^0 = \text{arg}\min_{W_{t}^0}  \Vert H_{t}^0W_{t}^0 -  ( Y - \alpha \sum_{t' < t} H_{t'}^0W_{t'}^0 )  \Vert^2 + \lambda \Vert W_{t}^0 \Vert^2, \quad \alpha \in (0,1),
\end{equation} 
\begin{equation}
W_{t}^0 = (H_{t}^{0T}H_{t}^0 + \lambda I)^{-1}H_{t}^{0T}(Y - \alpha \sum_{t' < t} H_{t'}^0W_{t'}^0), \quad t =1,...,T-1.
\end{equation} 
After $T$ iteration steps the first level will provide an approximation:
\begin{equation}
Y_0 = \alpha \sum_{t=0}^{T-1} H_{t}^0 W_{t}^0 = \alpha \sum_{t=0}^{T-1} h(XR^{0T}_t) W_{t}^0.
\end{equation}
Then we set:
\begin{equation}
Y'_0 = Y - Y_0,
\end{equation}
and we repeat the procedure for the next $L-1$ boosting levels, obtaining $Y_{\ell}$, $\ell=1,...,L-1$. 

The general equations for $\ell=1,...,L-1$ can be written as:
\begin{equation}
W_0^{\ell} = \text{arg}\min_{W_0^{\ell}}  \Vert H_0^{\ell}W_0^{\ell} -  Y'_{\ell-1}  \Vert^2 + \lambda \Vert W_0^{\ell} \Vert^2,
\end{equation}
\begin{equation}
W_0^{\ell}= (H_0^{\ell T}H_0^{\ell} + \lambda I)^{-1}H_0^{\ell T}Y'_{\ell-1},
\end{equation}
\begin{equation}
W_{t}^{\ell} = \text{arg}\min_{W_{t}^{\ell}}  \Vert H_{t}^{\ell}W_{t}^{\ell} -  Y'_{\ell-1} - \alpha \sum_{t' < t} H_{t'}^{\ell}W_{t'}^{\ell} )  \Vert^2 + \lambda \Vert W_{t}^{\ell} \Vert^2, \quad \alpha \in (0,1),
\end{equation} 
\begin{equation}
Y_{\ell} = \alpha \sum_{t=0}^{T-1} H_{t}^{\ell} W_{t}^{\ell} = \alpha \sum_{t=0}^{T-1} h(XR^{\ell T}_t) W_{t}^{\ell},
\end{equation}
\begin{equation}
Y'_{\ell} = Y - \sum_{\ell=0}^{L-1} Y_{\ell}.
\end{equation}

Given an unclassified data matrix $\tilde{X}=[\tilde{x}_{nm}]_{N'\times M}$, where each row is a new sample, we encode it using the same random projection matrices:
\begin{equation}
\tilde{H}_{\ell}^t=h(\tilde{X}R^T_{\ell}), \quad \ell=0,1,...,L-1, \quad t=0,1,...,T-1,
\end{equation}
we compute the output:
\begin{equation}
\tilde{Y} = \alpha \sum_{\ell=0}^{L-1} \sum_{t=0}^{T-1} \tilde{H}_{t}^{\ell} W_{t}^{\ell} = \alpha \sum_{\ell=0}^{L-1} \sum_{t=0}^{T-1} h(\tilde{X}R^{\ell T}_t) W_{t}^{\ell},
\end{equation}
and then we use the decision criterion (5) to decide the class for each new sample (row of $\tilde{X}$).

We should note that different random projection matrices $R_t^{\ell}$ are generated for each level and each time step: 
\begin{equation}
H_t^\ell = h(XR_t^{\ell T}), 
\end{equation}
and $\alpha \in (0,1)$ is a discount parameter (or a "learning" rate). Also, the random projection matrices can be generated on the fly, and they don't require additional storage. 

\section{Numerical results}

In order to illustrate the proposed method we use two well known data sets: MNIST \cite{key-6} and fashion-MNIST (fMNIST) \cite{key-7}. 
The MNIST data set is a large database of handwritten digits $\{0,1,...,9\}$, containing 60,000 training images 
and 10,000 testing images. These are monochrome images with an intensity in the interval $[0,255]$, and the size of $28 \times 28 = 784$ pixels. 
The fashion-MNIST dataset also consists of 60,000 training images and a test set of 10,000 images. The images are also monochrome, with an intensity in the interval $[0,255]$ and the size of $28 \times 28 = 784$ pixels. 
However, the fashion-MNIST is harder to classify, since it is a more complex dataset, containing images from $K=10$ different apparel classes: 
0 - t-shirt/top; 1 - trouser; 2 - pullover; 3 - dress; 4 - coat; 5 - sandal; 6 - shirt; 7 - sneaker; 8 - bag; 9 - ankle boot. 

In all numerical experiments we have used the following data normalization \cite{key-10}: 
\begin{equation}
x_n \leftarrow \sqrt{x_n}, \quad x_n \leftarrow x_n - \langle x_n \rangle, \quad x_n \leftarrow x_n/\Vert x_n \Vert.
\end{equation}

\subsection{tanh() activation}

In Figure 1 we give the classification accuracy for both data sets, as a function of the boosting level $\ell=0,1,...,19$, 
when the following parameters were kept fixed: $\mu=1$, $\alpha=1/2$, $T=50$, $J=M$. Also, in this case the activation function was $\text{tanh}()$, which is typically used in ELMs and other neural networks. 
One can see that the classification accuracy increases with the number of boosting levels $\ell$.  
In the case of MNIST the classification achieves an accuracy of $\eta > 98.9 \%$, for $\ell\geq 7$. 
Similarly in the case of fMINIST we obtain an accuracy of $\eta > 91.3 \%$, for $\ell\geq 7$. 
We should note that the regularization constant was set to a high value $\lambda=1$, which discourages overfitting. 

\subsection{sign() activation}

In a second experiment we replaced the tanh() activation function with the sign() function. This function is typically used in the approximate nearest neighbor problem, which can be stated as following \cite{key-8}, \cite{key-9}. 
Given a data set of points $X=\{x_0,x_1,...,x_{N-1}\}$, $x_n \in \mathbb{R}^M$, and a query point $q \in \mathbb{R}^M$, 
the goal is to find a point $x \in X$ such that $d(q,x) \leq (1+\varepsilon)d(q,x^*)$, $\varepsilon > 0$, where $x^*$ is the true nearest neighbor of $q$, and $d()$ is a distance measure. 
The similarity hashing maps the data set $X$ and the query $q$ to a set of hash values (hashes). 
In order to construct the hashes here we will use the random projection approach, 
which is known to preserve distance under the angular distance measures. Given an input vector $x\in \mathbb{R}^M$ and a random hyperplane defined by $r\in \mathbb{R}^M$, 
where $r_m \in \mathcal{N}(0,1)$ for $m=0,1,...,M-1$, we define a hashing function as:
\begin{equation}
h(x,r)=\text{sign}(x\cdot r) \in \{-1,+1\}, 
\end{equation}
where $\cdot$ is the dot product. 
With this choice of the hash function, one can easily show that given two vectors $x,x'\in \mathbb{R}^M$ the probability of $h(r,x)=h(r,x')$ is: 
\begin{equation}
\text{Pr}(h(x,r)=h(x',r)) = 1 - \frac{\theta(x,x')}{\pi}, 
\end{equation}
where $\theta(x,x')$ is the angle between $x$ and $x'$. 

We should note that each randomly drawn hyperplane defines a different hash function, and therefore to map the data vectors $x\in \mathbb{R}^M$ to $\{-1,+1\}^J$ we need to define $J$ hash functions:
\begin{equation}
h(r,x) = [h(x,r_0),h(x,r_1),...,h(x,r_{J-1})] \in \mathbb{R}^{J}.
\end{equation}
One can see that in our case, each column of the matrix $R=[r_{j,m}]_{J\times M}$ corresponds to a randomly drawn hyperplane. 

In Figure 2 we give the classification results. Once can see that method is quite robust, and the accuracy only drops by about $0.5\%$, comparing to the tanh() activation.  

\begin{figure}[!h]
\centering \includegraphics[width=10cm]{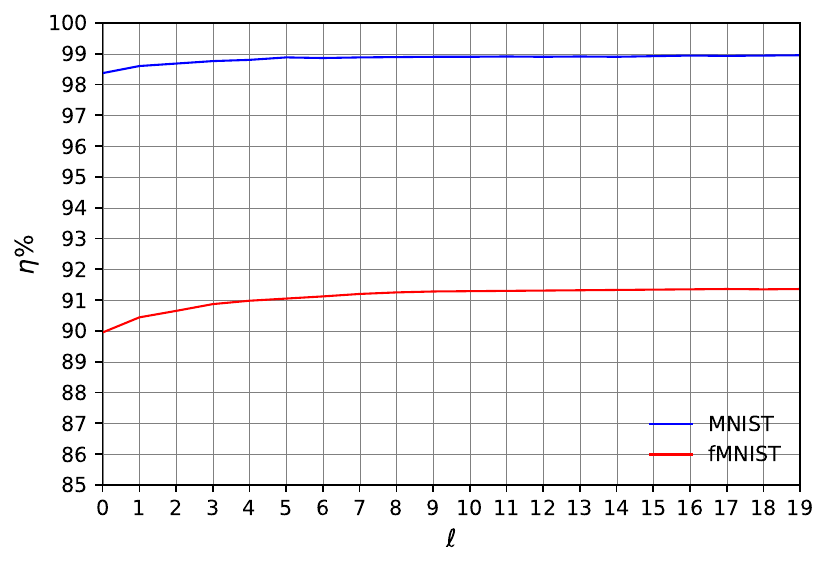}
\caption{The classification results for the tanh() activation function (see the text for details).}
\end{figure}

\begin{figure}[!h]
\centering \includegraphics[width=10cm]{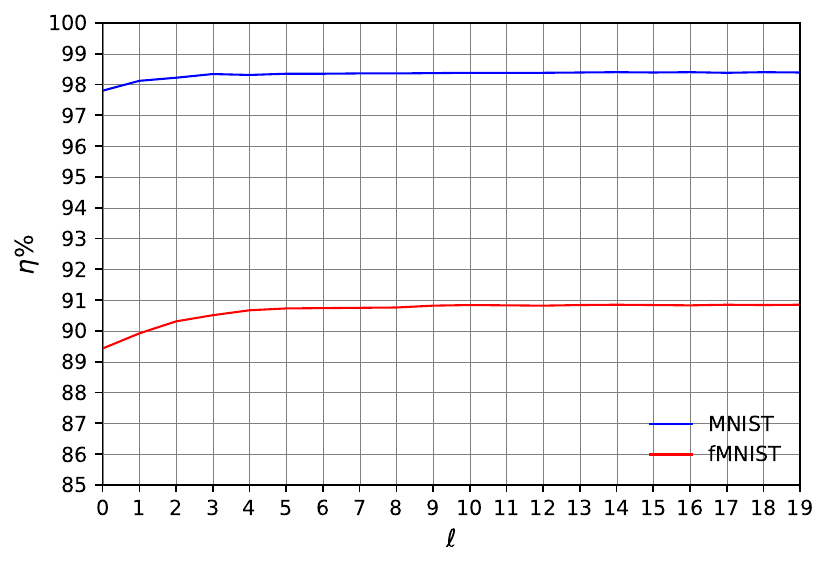}
\caption{The classification results for the sign() activation function (see the text for details).}
\end{figure}

\section*{Conclusion}

We have discussed a ridge regression boosting method with applications to ELMs. The proposed method significantly improves the accuracy of the ELM. 
The method is based "on the fly" random projections which do not require additional storage, only the random seed needs to be the same for both data training and testing processing. 
In the case of MNIST and fMNIST after about seven boosting levels the method saturates and no significant improvements can be seen. 
Besides the good classification results this boosting approach is also very robust to noise perturbations. 
For example, if $10\%$ of the pixels in the images are randomly set to zero, the classification 
accuracy is still reaching $98\%$ for MNIST and respectively $90\%$ for fMNIST.


\begin{thebibliography}{99}

\bibitem{key-1} 
P. F. Schmidt, M. A. Kraaijveld, R. P. W. Duin, \textit{Feed forward neural networks with random weights}, in Proc. 11th IAPR Int. Conf. on Pattern Recognition, Volume II, 
Conf. B: Pattern Recognition Methodology and Systems (ICPR11, The Hague, Aug.30 - Sep.3), IEEE Computer Society Press, Los Alamitos, CA, 1992, 1-4, 1992.

\bibitem{key-2} 
G.-B. Huang, Q.-Y. Zhu, C.-K. Siew, \textit{Extreme learning machine: Theory and applications}, Neurocomputing, 70(1-3) 489 (2006).

\bibitem{key-3} 
G. Tutz, H Binder, \textit{Boosting Ridge Regression}, Computational Statistics \& Data Analysis 51(12) 6044 (2007).

\bibitem{key-4} 
J. Bootkrajang, \textit{Boosting Ridge Regression for High Dimensional Data Classification}, arXiv:2003.11283 (2020).

\bibitem{key-5} 				
C. Peralez-Gonzalez, J. Perez-Rodriguez, A. M. Duran-Rosal, \textit{Boosting ridge for the extreme learning machine globally optimised for classification and regression problems}, Scientific Reports, 13, 11809 (2023).

\bibitem{key-6} 
Y. Lecun, L. Bottou, Y. Bengio, P. Haffner, \textit{Gradient-based learning applied to document recognition}, Proceedings of the IEEE 86(11), 2278 (1998).

\bibitem{key-7}
H. Xiao, K. Rasul, R. Vollgraf, \textit{Fashion-MNIST: a Novel Image Dataset for Benchmarking Machine Learning Algorithms}, arXiv:1708.07747 (2017).

\bibitem{key-8}
A. Gionis, P. Indyk, R. Motwani, \textit{Similarity Search in High Dimensions via Hashing}, VLDB '99: Proceedings of the 25th International Conference on Very Large Data Bases, 518 (1999)

\bibitem{key-9}
M. Charikar, \textit{Similarity Estimation Techniques from Rounding Algorithms}, STOC '02: Proceedings of the thiry-fourth annual ACM symposium on Theory of computing, 380 (2002).

\bibitem{key-10}
M. Andrecut, \textit{K-Means Kernel Classifier}, arXiv:2012.13021 (2020).

\end{thebibliography}
\end{document}